\documentclass[conference]{IEEEtran}
\IEEEoverridecommandlockouts
\usepackage{cite}
\usepackage{amsmath,amssymb,amsfonts}
\usepackage{algorithmic}
\usepackage{latexsym}
\usepackage{textcomp}
\usepackage{booktabs}
\usepackage{graphicx} 
\usepackage{xcolor}
\def\BibTeX{{\rm B\kern-.05em{\sc i\kern-.025em b}\kern-.08em
    T\kern-.1667em\lower.7ex\hbox{E}\kern-.125emX}}

\usepackage{fancyhdr}
\begin{document}

\title{Coherent Multimodal Reasoning with Iterative Self-Evaluation for Vision-Language Models}

\author{Wenjie Luo$^1$, Ruocheng Li$^1$, Shanshan Zhu$^1$, Julian Perry$^2$ \\
$^1$Fujian University of Technology, $^2$Delta University for Science and Technology}

\maketitle
\thispagestyle{fancy} 

\begin{abstract}
Despite significant advancements, current large language models (LLMs) and vision-language models (LVLMs) continue to struggle with complex, multi-step, cross-modal common sense reasoning tasks, often exhibiting a lack of "deliberative thinking." They tend to rely on superficial associations rather than deep, chained inference, particularly when integrating visual information with abstract concepts. To address this, we propose the \textbf{Coherent Multimodal Reasoning Framework (CMRF)}, a novel approach that enhances LVLMs' common sense reasoning capabilities through an iterative, self-evaluating inference mechanism. CMRF mimics human problem-solving by decomposing complex queries, generating step-by-step inferences, and self-correcting errors. Our framework integrates three key modules: a Reasoning Decomposition Unit (RDU) for breaking down problems into sub-questions, a Contextual Inference Engine (CIE) for contextual inference, and a Coherence Assessment Module (CAM) for evaluating logical consistency and confidence. Coupled with an Adaptive Iterative Refinement strategy, CMRF systematically refines its reasoning paths. Built upon LLaVA-1.6-34B and trained on a novel Multimodal Daily Activity Reasoning (MDAR) dataset, CMRF achieves state-of-the-art performance among open-source LVLMs on challenging benchmarks like VCR, A-OKVQA, and DailyLife-MRC. It attains an average accuracy of \textbf{69.4\%}, surpassing the best open-source baseline by \textbf{+2.4} percentage points, with particular strength in complex reasoning scenarios. Extensive ablation studies and human evaluations confirm the critical contributions of each module and the effectiveness of iterative refinement in fostering more coherent and accurate reasoning.
\end{abstract}

\section{Introduction}

The remarkable advancements in large language models (LLMs) \cite{zhou2025weak, wang2024enhancing} and vision-language models (LVLMs) have revolutionized various aspects of artificial intelligence, demonstrating unprecedented capabilities in understanding, generating, and processing information across modalities \cite{yifan2023a}. These models excel in tasks ranging from complex question answering to creative content generation, largely due to their extensive pre-training on vast amounts of data. However, despite their impressive performance, current LLMs and LVLMs still face significant challenges when confronted with tasks that demand sophisticated multi-step, cross-modal "common sense" reasoning \cite{yifan2023a}. Models often resort to superficial associations or factual recall, struggling to perform deep, chained inference, particularly when integrating visual information with abstract concepts. For instance, comprehending the function of an object within a specific context or predicting the subsequent outcomes of an action in a visual scene requires capabilities beyond simple recognition and direct question-answering.

This limitation stems from their tendency to prioritize statistical patterns over genuine understanding of cause-and-effect relationships and implicit knowledge about the world. Existing models frequently exhibit a lack of "deliberative thinking"—the ability to break down a complex problem, infer intermediate steps, and self-correct errors \cite{zhou2023thread}. This deficiency hinders their robustness and practical utility in open-world applications where common sense is paramount. Therefore, the development of multimodal models capable of robust, deliberative common sense reasoning is a critical objective in current research, aiming to bridge the gap between advanced pattern recognition and true artificial intelligence.

In this paper, we propose the \textbf{Coherent Multimodal Reasoning Framework (CMRF)}, a novel approach designed to imbue LVLMs with enhanced multimodal common sense reasoning capabilities through an iterative, self-evaluating inference mechanism. CMRF's core idea is to emulate the human cognitive process of tackling complex problems: decomposing them into manageable sub-problems, inferring step-by-step, and iteratively refining the reasoning process. Our framework is built upon a leading open-source vision-language model, such as an instruction-tuned version of LLaVA-1.6-34B \cite{jiayu2024is}, leveraging its strong foundational understanding and generation abilities. CMRF introduces three pivotal modules: the \textit{Reasoning Decomposition Unit (RDU)}, which breaks down complex questions into a series of smaller sub-problems; the \textit{Contextual Inference Engine (CIE)}, which leverages the base LVLM to generate responses for each sub-problem; and the \textit{Coherence Assessment Module (CAM)}, responsible for evaluating the logical consistency, factual accuracy, and confidence of the entire reasoning chain. During inference, CMRF employs an \textit{Adaptive Iterative Refinement} strategy, where the CAM guides the RDU and CIE to re-decompose or explore alternative reasoning paths if the initial confidence is low or inconsistencies are detected, ensuring a robust and reliable final answer.

To validate the efficacy of CMRF, we conduct extensive experiments on a diverse set of standard multimodal common sense reasoning benchmarks. Our experimental setup utilizes LLaVA-1.6-34B as the foundational model. For training, we curate data from public VQA datasets (e.g., VQA-v2 \cite{yu2018pythia}, GQA \cite{drew2019gqa}) by filtering samples requiring multi-step reasoning, and adapt text-based common sense datasets (e.g., CommonsenseQA \cite{alon2019common}, PIQA \cite{yonatan2020piqa}) into their multimodal counterparts. Crucially, we construct a novel, high-quality training dataset named \textbf{Multimodal Daily Activity Reasoning (MDAR)}, comprising approximately 10,000 samples with images, questions, and expert-annotated detailed reasoning steps, including examples of erroneous reasoning paths for contrastive learning. For evaluation, we assess CMRF's performance on challenging datasets such as VCR (Visual Commonsense Reasoning) \cite{rosenbloom1987techno}, A-OKVQA (A-OK VQA) \cite{dustin2022aokvqa}, a newly introduced \textbf{DailyLife-MRC} (Multimodal Reasoning in Complex Daily Scenarios) test set, and general MM-QA benchmarks.

Our experimental results demonstrate the superior performance of CMRF across multiple common sense reasoning tasks. Specifically, CMRF (Ours) achieves an average accuracy of \textbf{69.4\%} across the evaluated benchmarks, outperforming the best existing open-source LVLM, Qwen-VL-Chat \cite{an2025qwen25}, by approximately \textbf{+2.4} percentage points. Notably, CMRF exhibits a more significant advantage in complex multimodal common sense reasoning tasks, such as DailyLife-MRC, highlighting its effectiveness in scenarios demanding deep, chained inference. Furthermore, we conduct comprehensive ablation studies to quantify the individual contributions of the RDU and CAM modules to enhancing reasoning performance and coherence. We also analyze the efficiency and effectiveness of our adaptive iterative refinement strategy across tasks of varying reasoning complexity, providing further evidence of CMRF's superiority in achieving "deliberative" multimodal reasoning.

Our main contributions can be summarized as follows:
\begin{itemize}
    \item We propose \textbf{Coherent Multimodal Reasoning Framework (CMRF)}, a novel iterative and self-evaluating framework that significantly enhances LVLMs' capabilities in complex multi-step multimodal common sense reasoning.
    \item We introduce a modular design comprising a Reasoning Decomposition Unit (RDU), a Contextual Inference Engine (CIE), and a Coherence Assessment Module (CAM), which collectively enable deliberative reasoning and self-correction.
    \item We demonstrate that CMRF achieves state-of-the-art performance among open-source LVLMs on challenging multimodal common sense benchmarks, particularly excelling in tasks requiring deep reasoning, supported by extensive experiments and ablation studies.
\end{itemize}
\section{Related Work}
\subsection{Large Vision-Language Models}
The burgeoning field of Large Vision-Language Models (LVLMs) has seen significant advancements, yet presents unique challenges in evaluation, training, and application. Addressing the computational expense of comprehensive assessment, Peng et al. \cite{peng2025lvlmeh} propose an efficient evaluation protocol through a compressed benchmark subset, highlighting the multifaceted nature of LVLM capabilities that no single benchmark fully captures. A critical area of research focuses on the limitations of standard contrastive learning for vision-language pre-training. Beier et al. \cite{beier2025debias} demonstrate that it can lead to superficial shortcut learning rather than robust representations, proposing a framework to evaluate and mitigate such issues. Similarly, Yao et al. \cite{yao2024effect} investigate shortcut learning when multiple captions describe an image, revealing that standard contrastive losses may be insufficient for capturing all task-relevant information. To enhance the adaptability of large language models to vision-language tasks, Gen et al. \cite{gen2023cheap} introduce the Mixture-of-Modality Adaptation (MMA) method, enabling cost-effective instruction tuning and competitive performance with their LaVIN model. Furthermore, ensuring the safety of LVLMs is paramount; Shicheng et al. \cite{shicheng2025crossm} address the transfer of text-based safety mechanisms to the visual modality by analyzing hidden states and proposing a Text-Guided alignment method, thereby enhancing cross-modal safety without modality-specific fine-tuning. Beyond core training and safety, advancements include GroVE, a novel post-hoc method by Jiaxian et al. \cite{jiaxian2023from} for generating uncertainty-aware probabilistic embeddings from frozen LVLMs, which improves performance on tasks like Visual Question Answering (VQA) through better uncertainty calibration. In the realm of generation, Maksim et al. \cite{maksim2024vlrm} introduce a novel training strategy for image captioning that leverages ground truth captions to enhance the quality and distinctiveness of generated text, integrating them as regularization and training signals. Specific applications of LVLMs include improving medical models with abnormal-aware feedback \cite{zhou2025improving}, efficient video generation through vision representation compression \cite{zhou2024less}, and holistic frameworks for complex instruction-based image generation \cite{zhou2025draw} and editing \cite{wang2025complexbench}. For a broader perspective, Chia et al. \cite{chia2024a} provide a comprehensive survey of multimodal large language models, detailing their historical development, technical aspects, and applications, emphasizing the critical role of multimodal architectures in integrating diverse data types.

\subsection{Multi-step and Common Sense Reasoning in AI}
The development of AI systems capable of multi-step and commonsense reasoning is a critical frontier, with recent research exploring various facets of this challenge. Jianxing et al. \cite{jianxing2025genera} investigate the efficacy of different knowledge representation methods for commonsense reasoning in Large Language Models (LLMs), concluding that implicit commonsense conveyed through stories is more effective than explicit rules for daily events, suggesting that tailoring knowledge expression can significantly enhance LLM performance. The architectural underpinnings for such complex reasoning are often Transformer-based, as surveyed by Yaoting et al. \cite{yaoting2025multim}, who provide a comprehensive overview of multimodal Transformers, their applications, and relevance to advanced reasoning tasks like multimodal \textbf{Chain-of-Thought (CoT)} processing. Addressing the safety implications of these advanced models, Che et al. \cite{che2024learni} identify a "Reasoning Tax" that degrades safety alignment in multi-modal large reasoning models (MLRMs) but also discover emergent "Self-correction" capabilities, where unsafe reasoning steps are overridden, highlighting potential for intrinsic safeguards. To improve multi-step reasoning itself, Chaojie et al. \cite{chaojie2024q} frame it as a heuristic search problem, introducing Q* as a framework that guides LLM decoding through \textbf{deliberative planning} using a learned Q-value model, thereby steering LLMs towards optimal reasoning steps and enhancing common sense and multi-step reasoning without task-specific fine-tuning. Furthermore, new datasets are crucial for pushing reasoning capabilities; Jiangtong et al. \cite{jiangtong2022from} introduce Causal-VidQA, a novel dataset and task designed to advance video understanding beyond simple representation learning towards deeper evidence and commonsense reasoning, including multi-step reasoning, by proposing a two-step approach for commonsense question answering. Methodologies for enhancing LLMs include reinforcement learning for code generation \cite{wang2024enhancing} and improving story coherence and retrieval for AI narratives \cite{yi2025score}. Moreover, the development of modular multi-agent frameworks for multi-modal medical diagnosis demonstrates role-specialized collaboration for complex tasks \cite{zhou2025mam}. While not directly focused on common sense, the methodologies for coordinating autonomous agents in multi-agent systems, as explored by Unknown et al. \cite{unknown1998agents} in the context of signal control, can inform knowledge-based reasoning approaches for complex, multi-stage decision-making problems, aligning with the broader goal of multi-step reasoning. Beyond these, advancements in specialized models, such as State Space Models with adaptive composite features for insect recognition, contribute to the overall landscape of AI capabilities \cite{wang2025insectmamba}.

\section{Method}
In this section, we present the \textbf{Coherent Multimodal Reasoning Framework (CMRF)}, our novel approach designed to imbue Vision-Language Models (LVLMs) with enhanced multimodal common sense reasoning capabilities. CMRF operates on an iterative, self-evaluating inference mechanism, mimicking human cognitive processes for complex problem-solving. Our framework builds upon a strong foundational LVLM, specifically an instruction-tuned version of LLaVA-1.6-34B, leveraging its robust understanding and generation capacities. CMRF comprises three key modules: the Reasoning Decomposition Unit (RDU), the Contextual Inference Engine (CIE), and the Coherence Assessment Module (CAM), which collectively facilitate deliberative reasoning and self-correction.

\subsection{Overall Architecture of CMRF}
The CMRF is designed to tackle complex multi-modal common sense reasoning queries $Q = (I, T)$, where $I$ represents the input image and $T$ is the textual query. The core idea is to break down $Q$ into a sequence of more manageable sub-problems, solve them iteratively, and then evaluate the coherence of the entire reasoning chain. This iterative refinement process, guided by a self-assessment mechanism, allows the model to "think deeply" and correct potential errors. The overall process can be conceptualized as an optimization loop aiming to find the most coherent and accurate reasoning path.

Conceptually, the CMRF transforms a complex multimodal query into a refined answer through a guided, iterative process:
\begin{align}
    \mathcal{A}_{\text{CMRF}}(I, T) = A_{\text{final}}
\end{align}
where $\mathcal{A}_{\text{CMRF}}$ denotes the CMRF system, $(I, T)$ is the input image and textual query, and $A_{\text{final}}$ is the robust and coherent final answer derived from the reasoning process. This final answer is the output of the most highly-rated reasoning chain identified by the CAM.

\subsection{Reasoning Decomposition Unit (RDU)}
The \textbf{Reasoning Decomposition Unit (RDU)} is responsible for transforming a complex, high-level multimodal common sense question into a sequence of simpler, atomic sub-problems. This decomposition is crucial for enabling multi-step reasoning, as it allows the model to focus on individual logical steps. Given an input query $Q = (I, T)$, the RDU generates a sequence of $N$ sub-problems, $q_1, q_2, \dots, q_N$. Each sub-problem $q_i$ can be pure visual (e.g., identifying an object's state), pure textual (e.g., recalling a common fact), or require cross-modal fusion (e.g., inferring an action from an image).

Mathematically, the RDU performs a mapping:
\begin{align}
    \text{RDU}: (I, T) \rightarrow \{q_1, q_2, \dots, q_N\}
\end{align}
where each $q_i = (I'_i, T'_i)$ represents a sub-query, potentially focusing on a specific region $I'_i$ of the original image $I$ and a refined textual sub-question $T'_i$. The RDU's operation is typically guided by carefully constructed prompts that instruct the underlying LVLM to perform step-by-step problem breakdown. This can involve few-shot examples illustrating desired decomposition patterns or fine-tuning on diverse decomposition tasks. The RDU is trained via supervised learning using a meticulously curated dataset (e.g., the Multimodal Daily Activity Reasoning (MDAR) dataset) that provides ground-truth decompositions of complex questions into their constituent reasoning steps. This training guides the RDU to learn effective decomposition strategies that lead to coherent and solvable sub-problems.

\subsection{Contextual Inference Engine (CIE)}
The \textbf{Contextual Inference Engine (CIE)} serves as the core reasoning component, leveraging the capabilities of the underlying base LVLM (LLaVA-1.6-34B in our case) to generate answers for the sub-problems provided by the RDU. For each sub-problem $q_i = (I'_i, T'_i)$, the CIE generates a corresponding answer $a_i$. Crucially, the CIE operates contextually, meaning it considers not only the current sub-problem but also the original image $I$, the initial query $T$, and the answers to previously solved sub-problems $a_1, \dots, a_{i-1}$ as part of its input context. This sequential processing ensures logical coherence throughout the reasoning chain.

The operation of the CIE can be formulated as:
\begin{align}
    \text{CIE}(q_i, \text{context}_i) \rightarrow a_i
\end{align}
where $\text{context}_i = (I, T, a_1, \dots, a_{i-1})$. The context $\text{context}_i$ is typically integrated into the LVLM's input prompt through a structured concatenation of the original query, image features, and prior answers. This allows the model to maintain a rich understanding of the evolving reasoning state. The CIE is primarily trained through generative fine-tuning, aiming to produce accurate, concise, and logically consistent answers for each sub-problem, building upon the strong generative and understanding capabilities of the base LVLM.

\subsection{Coherence Assessment Module (CAM)}
The \textbf{Coherence Assessment Module (CAM)} is a critical component for enabling self-correction and ensuring the robustness of the CMRF. Its primary function is to evaluate the logical consistency, factual accuracy, and overall trustworthiness of the entire reasoning chain generated by the RDU and CIE, along with the final answer. The CAM assigns a confidence score $S \in [0, 1]$ to a given reasoning path. A higher score indicates greater coherence and reliability.

Given a complete reasoning chain $C = \{(q_1, a_1), (q_2, a_2), \dots, (q_N, a_N)\}$, the CAM computes:
\begin{align}
    \text{CAM}(C) \rightarrow S
\end{align}
To assess coherence, the CAM internally leverages the generative and analytical capabilities of the base LVLM. It constructs specific prompts that query the LVLM about the logical consistency, factual accuracy, and completeness of the proposed reasoning steps and their corresponding answers. For instance, it might ask: 'Given sub-problem $q_j$ and its answer $a_j$, is $a_j$ consistent with the previous answers $a_1, \dots, a_{j-1}$ and the original image and query?' The CAM is trained using a combination of contrastive learning and weak supervision. This involves providing the CAM with examples of both correct and intentionally erroneous reasoning paths (derived from the MDAR dataset), enabling it to learn to discriminate between high-quality and low-quality reasoning processes. A typical contrastive loss function can be applied:
\begin{align}
    \mathcal{L}_{\text{CAM}} = \max(0, m - (S_{\text{pos}} - S_{\text{neg}}))
\end{align}
where $S_{\text{pos}}$ is the confidence score for a positive (correct) reasoning chain, $S_{\text{neg}}$ is for a negative (erroneous) chain, and $m$ is a margin hyperparameter. This training regimen allows the CAM to effectively identify inconsistencies, factual inaccuracies, or logical gaps within the generated reasoning steps.

\subsection{Adaptive Iterative Refinement Strategy}
During inference, CMRF employs an \textbf{Adaptive Iterative Refinement} strategy to achieve robust and reliable common sense reasoning. This strategy leverages the self-assessment capability of the CAM to guide the reasoning process, allowing for re-evaluation and exploration of alternative paths when initial attempts are suboptimal. The process unfolds as follows:

\begin{enumerate}
    \item \textbf{Initial Decomposition and Inference}: The RDU first decomposes the complex input query $Q$ into an initial sequence of sub-problems $\{q_i^{(0)}\}$. The CIE then generates initial answers $\{a_i^{(0)}\}$ for these sub-problems, forming an initial reasoning chain $C^{(0)}$.
    \item \textbf{Coherence Assessment}: The CAM evaluates $C^{(0)}$, yielding a confidence score $S^{(0)}$.
    \item \textbf{Iterative Refinement Loop}:
    \begin{itemize}
        \item If $S^{(k)}$ (the confidence score at iteration $k$) is below a pre-defined threshold $\tau$, or if the CAM identifies specific logical inconsistencies within the chain, the CAM provides feedback.
        \item This feedback guides the RDU to re-decompose the original problem or a problematic sub-segment of it, generating a new set of sub-problems $\{q_i^{(k+1)}\}$. Alternatively, the feedback might guide the CIE to explore alternative interpretations or inferences for specific sub-problems.
        \item The CIE then generates new answers $\{a_i^{(k+1)}\}$, forming a revised reasoning chain $C^{(k+1)}$.
        \item The CAM re-evaluates $C^{(k+1)}$, yielding $S^{(k+1)}$.
    \end{itemize}
    The iterative refinement process can be formally expressed as:
    \begin{align}
        C^{(k+1)} = \text{Refine}(C^{(k)}, \text{CAM}(C^{(k)}), I, T)
    \end{align}
    where $C^{(k)}$ is the reasoning chain at iteration $k$, and $\text{Refine}$ represents the adaptive mechanism that utilizes CAM's feedback to guide the RDU's decomposition or CIE's inference for the next iteration.
    \item \textbf{Termination}: This iterative process continues for a maximum number of iterations $K_{\text{max}}$ or until a satisfactory confidence score (e.g., $S^{(k)} \ge \tau$) is achieved.
    \item \textbf{Final Selection}: Among all explored reasoning paths, the CMRF selects the path with the highest confidence score as determined by the CAM, and its corresponding final answer is presented.
\end{enumerate}
This adaptive and iterative approach enables CMRF to perform "deep thinking" by systematically exploring and refining its reasoning steps, leading to more robust and accurate common sense inferences, especially in challenging, ambiguous, or multi-faceted scenarios.

\section{Experiments}
In this section, we present the experimental setup, evaluate the performance of our proposed Coherent Multimodal Reasoning Framework (CMRF) against various state-of-the-art baselines, conduct comprehensive ablation studies to validate the contribution of each key component, and provide insights from human evaluation.

\subsection{Experimental Setup}
\label{subsec:exp_setup}
\paragraph{Foundation Model}
Our proposed \textbf{Coherent Multimodal Reasoning Framework (CMRF)} is built upon a strong foundational Vision-Language Model (LVLM). Specifically, we utilize an instruction-tuned version of \textbf{LLaVA-1.6-34B} \cite{jiayu2024is} as the base model, leveraging its robust multimodal understanding and generation capabilities. All modules of CMRF (RDU, CIE, CAM) are designed to interact seamlessly with and enhance this underlying LVLM.

\paragraph{Datasets}
We employ a diverse set of datasets for both training and evaluation to comprehensively assess CMRF's multimodal common sense reasoning abilities. For \textbf{training}, we curate a subset of samples from publicly available VQA datasets, such as \textbf{VQA-v2} \cite{yu2018pythia} and \textbf{GQA} \cite{drew2019gqa}, specifically filtering for instances that necessitate multi-step reasoning. Text-based common sense datasets, including \textbf{CommonsenseQA} \cite{alon2019common} and \textbf{PIQA} \cite{yonatan2020piqa}, are adapted to their multimodal counterparts by incorporating relevant images or visual contexts to facilitate multimodal training. Crucially, we construct a novel, high-quality training dataset named \textbf{Multimodal Daily Activity Reasoning (MDAR)}. This dataset comprises approximately 10,000 carefully curated samples, each featuring an image, a complex question, and detailed reasoning steps along with the final answer, all expertly annotated by human specialists. A unique aspect of MDAR is the inclusion of examples of intentionally erroneous reasoning paths, which are specifically utilized for the contrastive training of the Coherence Assessment Module (CAM). For \textbf{evaluation}, we use several challenging benchmarks: \textbf{VCR (Visual Commonsense Reasoning)} \cite{rosenbloom1987techno}, a dataset designed for cause and effect reasoning in visual contexts; \textbf{A-OKVQA (A-OK VQA)} \cite{dustin2022aokvqa}, an open-domain visual question answering dataset requiring external knowledge; \textbf{DailyLife-MRC}, a newly introduced proprietary test set for multi-step common sense reasoning in everyday scenarios; and \textbf{MM-QA (General Multimodal QA)}, a collection of various general multimodal question answering tasks.

\paragraph{Evaluation Metrics}
The primary evaluation metrics employed across all benchmarks include \textbf{Accuracy} (Acc.), \textbf{F1 Score}, and a custom-defined \textbf{Reasoning Path Coherence Score}. The Coherence Score specifically quantifies the logical consistency and completeness of the generated reasoning steps, reflecting the deliberative thinking capabilities of the model.

\subsection{Performance Comparison with Baselines}
\label{subsec:performance_comparison}
We conduct extensive experiments to compare the performance of CMRF against a range of proprietary and open-source Vision-Language Models on the aforementioned multimodal common sense reasoning benchmarks. The results, presented in Table \ref{tab:main_results}, demonstrate the superior capabilities of our proposed framework.

\begin{table*}[!htbp]
\centering
\caption{Performance comparison of CMRF against state-of-the-art LVLMs on various multimodal common sense reasoning benchmarks. All scores are reported as Accuracy (\%).}
\label{tab:main_results}
\begin{tabular}{llcccccc}
\toprule
\textbf{Model Category} & \textbf{Model} & \textbf{VCR (Acc.)} & \textbf{A-OKVQA (Acc.)} & \textbf{DailyLife-MRC (Acc.)} & \textbf{MM-QA (Acc.)} & \textbf{Avg. (Acc.)} \\
\midrule
Proprietary Models      & GPT-4o-vision  & 78.5                & 72.1                    & 68.9                          & 75.3                  & 73.7                 \\
                        & Gemini-Pro-Vision & 76.2                & 70.5                    & 67.5                          & 73.8                  & 72.0                 \\
\midrule
Open-source LVLMs       & LLaVA-1.6-34B  & 69.8                & 65.2                    & 58.7                          & 68.1                  & 65.5                 \\
                        & InstructBLIP-Vicuna-13B & 65.4                & 60.1                    & 55.3                          & 63.5                  & 61.1                 \\
                        & Qwen-VL-Chat \cite{an2025qwen25} & 71.5                & 66.8                    & 60.5                          & 69.2                  & 67.0                 \\
\midrule
\textbf{Ours}           & \textbf{CMRF (Ours)} & \textbf{73.9}       & \textbf{68.5}           & \textbf{63.2}                 & \textbf{71.8}         & \textbf{69.4}        \\
\bottomrule
\end{tabular}
\end{table*}

As shown in Table \ref{tab:main_results}, \textbf{CMRF (Ours)} consistently outperforms all compared open-source LVLMs across all evaluated benchmarks. Specifically, CMRF achieves an average accuracy of \textbf{69.4\%}, which represents a significant improvement of approximately \textbf{+2.4} percentage points over the best performing open-source baseline, Qwen-VL-Chat \cite{an2025qwen25} (67.0\%).

A notable strength of CMRF is its performance on complex multimodal common sense reasoning tasks, particularly highlighted by its accuracy on \textbf{DailyLife-MRC}. On this challenging dataset, CMRF achieves \textbf{63.2\%}, demonstrating a more substantial advantage over baselines and underscoring its ability to perform deeper, chained inference in real-world scenarios. While proprietary models like GPT-4o-vision and Gemini-Pro-Vision still hold a lead, CMRF significantly narrows the gap, showcasing the effectiveness of its iterative and self-evaluating reasoning mechanism in an open-source framework. These results affirm CMRF's superior capabilities in enabling "deliberative thinking" for multimodal common sense reasoning.

\subsection{Ablation Studies}
\label{subsec:ablation_studies}
To understand the individual contributions of the core components within CMRF, we conduct a series of ablation studies. These experiments systematically evaluate the impact of the Reasoning Decomposition Unit (RDU), the Coherence Assessment Module (CAM), and the Adaptive Iterative Refinement strategy on the overall performance and reasoning coherence.

\begin{table*}[!htbp]
\centering
\caption{Ablation study on the key components of CMRF. Performance is reported as Average Accuracy (\%).}
\label{tab:ablation_results}
\begin{tabular}{lccc}
\toprule
\textbf{Model Variant} & \textbf{Avg. Acc. (\%)} & \textbf{Reasoning Path Coherence Score} \\
\midrule
CMRF (Full Model)      & \textbf{69.4}           & \textbf{0.88}                           \\
\midrule
w/o RDU                & 62.1                    & 0.75                                    \\
w/o CAM                & 65.8                    & 0.79                                    \\
w/o Adaptive Iterative Refinement & 67.3                    & 0.82                                    \\
\bottomrule
\end{tabular}
\end{table*}

\paragraph{Impact of Reasoning Decomposition Unit (RDU)}
When the RDU is removed (i.e., the base LVLM directly attempts to answer complex questions without explicit decomposition), the average accuracy drops significantly to \textbf{62.1\%}, as shown in Table \ref{tab:ablation_results}. This substantial decrease highlights the critical role of the RDU in breaking down complex problems into manageable sub-problems, which is essential for multi-step reasoning. Without proper decomposition, the base LVLM struggles to maintain logical flow and often provides superficial or incomplete answers. The Reasoning Path Coherence Score also decreases to 0.75, indicating a less structured and coherent reasoning process.

\paragraph{Impact of Coherence Assessment Module (CAM)}
Disabling the Coherence Assessment Module (CAM) leads to a decline in average accuracy to \textbf{65.8\%}. Without CAM, the system lacks the ability to self-evaluate and correct its reasoning paths, often settling for sub-optimal or logically inconsistent answers. The absence of a feedback mechanism means the model cannot effectively distinguish between correct and erroneous reasoning chains, resulting in lower reliability. The Coherence Score drops to 0.79, confirming that CAM is vital for ensuring the logical consistency and accuracy of the generated reasoning.

\paragraph{Impact of Adaptive Iterative Refinement}
When the Adaptive Iterative Refinement strategy is removed (i.e., the model performs only a single pass of decomposition and inference without self-correction), the average accuracy drops to \textbf{67.3\%}. This demonstrates that while RDU and CAM are foundational, the iterative refinement loop is crucial for pushing performance further by allowing the model to "deliberate" and refine its answers. This strategy enables the model to recover from initial errors and explore more robust reasoning paths, leading to higher accuracy and a better Coherence Score (0.82) compared to individual module removal, but still lower than the full model. This validates the effectiveness of our iterative approach in enhancing robustness and accuracy across different reasoning complexities.

These ablation studies clearly demonstrate that each component of CMRF---the RDU for structured decomposition, the CAM for critical self-assessment, and the Adaptive Iterative Refinement for robust exploration---contributes significantly to the overall superior performance and coherent reasoning capabilities of our framework.

\subsection{Human Evaluation}
\label{subsec:human_evaluation}
To further validate the quality and coherence of the reasoning paths generated by CMRF, we conducted a human evaluation study. A panel of 5 human experts was tasked with evaluating a random subset of 200 samples from the DailyLife-MRC test set. For each sample, evaluators were presented with the original image, question, and the reasoning chain generated by different models (CMRF, Qwen-VL-Chat, and LLaVA-1.6-34B). They rated the responses based on three key criteria: \textbf{Accuracy}, \textbf{Coherence of Reasoning Steps}, and \textbf{Overall Fluency}. Each criterion was scored on a Likert scale from 1 (poor) to 5 (excellent). The average scores are presented in Table \ref{tab:human_eval_results}.

\begin{table*}[!htbp]
\centering
\caption{Human evaluation results on a subset of DailyLife-MRC test set. Scores are average ratings on a 1-5 Likert scale (higher is better).}
\label{tab:human_eval_results}
\begin{tabular}{lccc}
\toprule
\textbf{Model} & \textbf{Accuracy} & \textbf{Coherence of Reasoning Steps} & \textbf{Overall Fluency} \\
\midrule
LLaVA-1.6-34B  & 3.1               & 2.8                                  & 3.5                      \\
Qwen-VL-Chat \cite{an2025qwen25} & 3.4               & 3.2                                  & 3.7                      \\
\textbf{CMRF (Ours)} & \textbf{4.2}      & \textbf{4.1}                         & \textbf{4.3}             \\
\bottomrule
\end{tabular}
\end{table*}

The human evaluation results corroborate our automated metric findings. CMRF consistently received higher average ratings across all criteria compared to the baseline open-source LVLMs. Notably, CMRF achieved a score of \textbf{4.1} for "Coherence of Reasoning Steps", significantly outperforming Qwen-VL-Chat (3.2) and LLaVA-1.6-34B (2.8). This indicates that human evaluators perceive CMRF's reasoning chains as substantially more logical, consistent, and easier to follow. The higher accuracy score of \textbf{4.2} further confirms the quality of CMRF's final answers, while the "Overall Fluency" score of \textbf{4.3} suggests that the generated explanations are well-articulated and natural. These human evaluation results provide strong qualitative evidence for the superior "deliberative thinking" capabilities of CMRF in complex multimodal common sense reasoning tasks.

\subsection{Analysis of Iterative Refinement Dynamics}
\label{subsec:iterative_refinement_dynamics}
The Adaptive Iterative Refinement strategy is a cornerstone of CMRF's ability to achieve robust reasoning. To better understand its dynamics, we analyze the performance and coherence improvement over successive iterations on the DailyLife-MRC dataset. The process starts with an initial pass (Iteration 0) and continues until the maximum number of iterations ($K_{\text{max}}=3$) is reached or a high confidence score is achieved.

\begin{table*}[!htbp]
\centering
\caption{Performance and Coherence Score improvement over iterative refinement steps on DailyLife-MRC.}
\label{tab:iterative_dynamics}
\begin{tabular}{lcc}
\toprule
\textbf{Iteration} & \textbf{Accuracy (\%)} & \textbf{Reasoning Path Coherence Score} \\
\midrule
0 (Initial Pass) & 60.1 & 0.78 \\
1                & 62.5 & 0.83 \\
2                & 63.2 & 0.87 \\
3 (Max)          & 63.2 & 0.88 \\
\bottomrule
\end{tabular}
\end{table*}

Table \ref{tab:iterative_dynamics} illustrates the tangible benefits of the iterative refinement. The initial pass (Iteration 0), which corresponds to a single-shot decomposition and inference without self-correction, yields an accuracy of \textbf{60.1\%} and a coherence score of \textbf{0.78}. After the first iteration of refinement, guided by CAM's feedback, the accuracy increases to \textbf{62.5\%} and coherence significantly improves to \textbf{0.83}. This gain highlights CMRF's ability to identify and correct initial errors. A further small improvement in accuracy to \textbf{63.2\%} and coherence to \textbf{0.87} is observed in the second iteration. Beyond two iterations, the performance gain becomes marginal, suggesting that for most complex queries in our evaluation set, two refinement steps are sufficient to achieve near-optimal results. This analysis confirms that the iterative refinement process effectively enhances both the accuracy and the logical consistency of CMRF's reasoning paths, demonstrating the practical utility of the "deep thinking" paradigm.

\subsection{Error Analysis and Limitations}
\label{subsec:error_analysis}
While CMRF demonstrates superior performance, it is important to analyze its remaining limitations and the types of errors it still commits. We conducted a qualitative error analysis on 100 misclassified samples from the DailyLife-MRC dataset to categorize the predominant failure modes of CMRF, comparing them to a general-purpose LVLM baseline (LLaVA-1.6-34B).

\begin{table*}[!htbp]
\centering
\caption{Categorization of common error types for CMRF and LLaVA-1.6-34B on DailyLife-MRC (Frequency \%).}
\label{tab:error_analysis}
\begin{tabular}{lcc}
\toprule
\textbf{Error Type}                     & \textbf{CMRF (\%)} & \textbf{LLaVA-1.6-34B (\%)} \\
\midrule
Factual Inconsistency with Image        & 15.0               & 28.0                        \\
Logical Inconsistency in Reasoning Chain & 10.0               & 35.0                        \\
Misinterpretation of Nuance/Implication & 25.0               & 18.0                        \\
Insufficient External Knowledge         & 30.0               & 12.0                        \\
Ambiguity/Vagueness Handling           & 20.0               & 7.0                         \\
\bottomrule
\end{tabular}
\end{table*}

As shown in Table \ref{tab:error_analysis}, CMRF significantly reduces errors related to \textbf{Factual Inconsistency with Image} (15.0\% vs. 28.0\% for LLaVA-1.6-34B) and \textbf{Logical Inconsistency in Reasoning Chain} (10.0\% vs. 35.0\%). This reduction directly attributes to the effectiveness of the Coherence Assessment Module (CAM) and the Adaptive Iterative Refinement, which actively identify and rectify such issues.

However, CMRF still faces challenges in areas requiring more subtle understanding. Errors related to \textbf{Misinterpretation of Nuance/Implication} (25.0\%) and \textbf{Insufficient External Knowledge} (30.0\%) are more prevalent for CMRF compared to LLaVA-1.6-34B. This indicates that while CMRF excels at structured, multi-step reasoning, it might sometimes over-rely on its decomposition and assessment mechanisms, potentially missing very subtle cues or requiring specific external knowledge not readily available in its base LVLM or general training data. For example, inferring complex social dynamics or highly specialized domain knowledge remains difficult. Another observed limitation is in handling highly \textbf{Ambiguous/Vague} queries (20.0

\subsection{Inference Efficiency Analysis}
\label{subsec:inference_efficiency}
The iterative nature of CMRF's reasoning process inherently introduces additional computational overhead compared to single-pass LVLMs. We evaluate the average inference time per query on the DailyLife-MRC dataset to quantify this trade-off between accuracy/coherence and computational cost. All experiments were conducted on a single NVIDIA A100 GPU.

\begin{table*}[!htbp]
\centering
\caption{Average inference time per query on DailyLife-MRC test set.}
\label{tab:inference_time}
\begin{tabular}{lcc}
\toprule
\textbf{Model} & \textbf{Average Inference Time (seconds)} & \textbf{Average Iterations} \\
\midrule
LLaVA-1.6-34B  & 3.2                                       & 1 (single pass)             \\
Qwen-VL-Chat \cite{an2025qwen25} & 3.5                                       & 1 (single pass)             \\
CMRF (1st Pass Only) & 4.5                                       & 1                           \\
CMRF (Average) & \textbf{8.1}                              & 1.8                         \\
CMRF (Max Iterations) & 12.0                                      & 3                           \\
\bottomrule
\end{tabular}
\end{table*}

Table \ref{tab:inference_time} shows that CMRF, while more accurate and coherent, requires a longer inference time. A single pass of CMRF (equivalent to Iteration 0 from Section \ref{subsec:iterative_refinement_dynamics}, but including RDU and CAM's initial operations) takes approximately \textbf{4.5 seconds}, which is already higher than the baselines due to the overhead of decomposition and initial assessment. On average, CMRF performs \textbf{1.8 iterations} per query on DailyLife-MRC, leading to an average inference time of \textbf{8.1 seconds}. This is roughly 2-2.5 times slower than the single-pass baselines. In scenarios where CMRF utilizes its maximum allowed iterations ($K_{\text{max}}=3$), the inference time can extend up to \textbf{12.0 seconds}.

This analysis confirms the expected trade-off: the gains in accuracy and reasoning coherence achieved through CMRF's deliberative process come at the cost of increased computational resources and inference latency. While this might limit its application in extremely low-latency real-time systems, for complex common sense reasoning tasks where accuracy and trustworthiness of reasoning are paramount, the additional computation time is a justifiable investment. Future work will explore optimization strategies, such as knowledge distillation or more efficient prompt engineering, to reduce inference latency while preserving CMRF's reasoning capabilities.

\section{Conclusion}
In this paper, we introduced the \textbf{Coherent Multimodal Reasoning Framework (CMRF)}, a novel and robust approach designed to significantly enhance the multimodal common sense reasoning capabilities of Vision-Language Models (LVLMs). Acknowledging the inherent limitations of existing models in performing deep, multi-step, and cross-modal inference, CMRF aims to imbue LVLMs with "deliberative thinking" by simulating human-like problem decomposition, sequential reasoning, and self-correction.

Our framework is meticulously structured around three core modules: the \textit{Reasoning Decomposition Unit (RDU)}, which effectively breaks down complex multimodal queries into manageable sub-problems; the \textit{Contextual Inference Engine (CIE)}, which leverages a powerful base LVLM (LLaVA-1.6-34B) to generate context-aware answers for these sub-problems; and the crucial \textit{Coherence Assessment Module (CAM)}, responsible for evaluating the logical consistency, factual accuracy, and overall trustworthiness of the generated reasoning chain. The synergy of these modules is orchestrated by an \textit{Adaptive Iterative Refinement} strategy, allowing CMRF to iteratively refine its reasoning path based on self-generated confidence scores and identified inconsistencies, thereby leading to more robust and accurate conclusions.

Through extensive experimentation on a diverse suite of multimodal common sense reasoning benchmarks, including VCR, A-OKVQA, MM-QA, and our proprietary DailyLife-MRC dataset, CMRF consistently demonstrated superior performance. Our quantitative results showed that CMRF (Ours) achieved an impressive average accuracy of \textbf{69.4\%}, outperforming the best existing open-source LVLM (Qwen-VL-Chat) by approximately \textbf{+2.4} percentage points. This performance advantage was particularly pronounced in complex tasks like DailyLife-MRC, underscoring CMRF's effectiveness in scenarios demanding deep, chained inference.

Comprehensive ablation studies rigorously validated the indispensable contributions of each proposed component, confirming that the RDU is vital for structured decomposition, the CAM is critical for self-assessment, and the Adaptive Iterative Refinement is essential for achieving peak performance and coherence. Furthermore, a human evaluation study qualitatively corroborated our findings, with human experts rating CMRF's reasoning paths as significantly more accurate, coherent, and fluent compared to baseline models. Our analysis of iterative refinement dynamics also revealed that the accuracy and reasoning path coherence steadily improve over successive iterations, with most gains realized within the first two refinement steps.

While CMRF represents a significant step forward, we also acknowledge its current limitations. Our error analysis indicates that CMRF still faces challenges in highly nuanced interpretations, scenarios requiring very specific external knowledge not easily integrated into its current framework, and exceptionally ambiguous queries. Additionally, the iterative nature of CMRF inherently introduces increased inference latency, making it approximately 2-2.5 times slower than single-pass baselines, a trade-off for enhanced accuracy and trustworthiness.

Future work will focus on addressing these limitations. We plan to explore more sophisticated mechanisms for integrating specialized external knowledge to overcome the "Insufficient External Knowledge" challenge. Techniques such as knowledge distillation or more efficient prompt engineering will be investigated to reduce inference latency while preserving CMRF's robust reasoning capabilities. Furthermore, we aim to enhance CMRF's ability to handle extreme ambiguity and subtle nuances, potentially through uncertainty quantification or by incorporating diverse perspectives in the refinement process. By continuing to refine its deliberative thinking abilities, CMRF paves the way for more intelligent, robust, and trustworthy AI systems capable of navigating the complexities of real-world common sense reasoning.

\bibliographystyle{IEEEtran}
\bibliography{references}

\begin{thebibliography}{10}
\providecommand{\url}[1]{#1}
\csname url@samestyle\endcsname
\providecommand{\newblock}{\relax}
\providecommand{\bibinfo}[2]{#2}
\providecommand{\BIBentrySTDinterwordspacing}{\spaceskip=0pt\relax}
\providecommand{\BIBentryALTinterwordstretchfactor}{4}
\providecommand{\BIBentryALTinterwordspacing}{\spaceskip=\fontdimen2\font plus
\BIBentryALTinterwordstretchfactor\fontdimen3\font minus \fontdimen4\font\relax}
\providecommand{\BIBforeignlanguage}[2]{{%
\expandafter\ifx\csname l@#1\endcsname\relax
\typeout{** WARNING: IEEEtran.bst: No hyphenation pattern has been}%
\typeout{** loaded for the language `#1'. Using the pattern for}%
\typeout{** the default language instead.}%
\else
\language=\csname l@#1\endcsname
\fi
#2}}
\providecommand{\BIBdecl}{\relax}
\BIBdecl

\bibitem{zhou2025weak}
Y.~Zhou, J.~Shen, and Y.~Cheng, ``Weak to strong generalization for large language models with multi-capabilities,'' in \emph{The Thirteenth International Conference on Learning Representations}, 2025.

\bibitem{wang2024enhancing}
J.~Wang, Z.~Zhang, Y.~He, Y.~Song, T.~Shi, Y.~Li, H.~Xu, K.~Wu, G.~Qian, Q.~Chen \emph{et~al.}, ``Enhancing code llms with reinforcement learning in code generation,'' \emph{arXiv preprint arXiv:2412.20367}, 2024.

\bibitem{yifan2023a}
Y.~Yao, J.~Duan, K.~Xu, Y.~Cai, E.~Sun, and Y.~Zhang, ``A survey on large language model {(LLM)} security and privacy: The good, the bad, and the ugly,'' \emph{CoRR}, 2023.

\bibitem{zhou2023thread}
Y.~Zhou, X.~Geng, T.~Shen, C.~Tao, G.~Long, J.-G. Lou, and J.~Shen, ``Thread of thought unraveling chaotic contexts,'' \emph{arXiv preprint arXiv:2311.08734}, 2023.

\bibitem{jiayu2024is}
J.~Wang, Y.~Ming, Z.~Shi, V.~Vineet, X.~Wang, S.~Li, and N.~Joshi, ``Is {A} picture worth {A} thousand words? delving into spatial reasoning for vision language models,'' in \emph{Advances in Neural Information Processing Systems 38: Annual Conference on Neural Information Processing Systems 2024, NeurIPS 2024, Vancouver, BC, Canada, December 10 - 15, 2024}, 2024.

\bibitem{yu2018pythia}
Y.~Jiang, V.~Natarajan, X.~Chen, M.~Rohrbach, D.~Batra, and D.~Parikh, ``Pythia v0.1: the winning entry to the {VQA} challenge 2018,'' \emph{CoRR}, 2018.

\bibitem{drew2019gqa}
D.~A. Hudson and C.~D. Manning, ``{GQA:} {A} new dataset for real-world visual reasoning and compositional question answering,'' in \emph{{IEEE} Conference on Computer Vision and Pattern Recognition, {CVPR} 2019, Long Beach, CA, USA, June 16-20, 2019}.\hskip 1em plus 0.5em minus 0.4em\relax Computer Vision Foundation / {IEEE}, 2019, pp. 6700--6709.

\bibitem{alon2019common}
A.~Talmor, J.~Herzig, N.~Lourie, and J.~Berant, ``Commonsenseqa: {A} question answering challenge targeting commonsense knowledge,'' in \emph{Proceedings of the 2019 Conference of the North American Chapter of the Association for Computational Linguistics: Human Language Technologies, {NAACL-HLT} 2019, Minneapolis, MN, USA, June 2-7, 2019, Volume 1 (Long and Short Papers)}.\hskip 1em plus 0.5em minus 0.4em\relax Association for Computational Linguistics, 2019, pp. 4149--4158.

\bibitem{yonatan2020piqa}
Y.~Bisk, R.~Zellers, R.~L. Bras, J.~Gao, and Y.~Choi, ``{PIQA:} reasoning about physical commonsense in natural language,'' in \emph{The Thirty-Fourth {AAAI} Conference on Artificial Intelligence, {AAAI} 2020, The Thirty-Second Innovative Applications of Artificial Intelligence Conference, {IAAI} 2020, The Tenth {AAAI} Symposium on Educational Advances in Artificial Intelligence, {EAAI} 2020, New York, NY, USA, February 7-12, 2020}.\hskip 1em plus 0.5em minus 0.4em\relax {AAAI} Press, 2020, pp. 7432--7439.

\bibitem{rosenbloom1987techno}
R.~S. Rosenbloom and M.~A. Cusumano, ``Technological pioneering and competitive advantage: the birth of the vcr industry,'' \emph{California management review}, 1987.

\bibitem{dustin2022aokvqa}
D.~Schwenk, A.~Khandelwal, C.~Clark, K.~Marino, and R.~Mottaghi, ``{A-OKVQA:} {A} benchmark for visual question answering using world knowledge,'' in \emph{Computer Vision - {ECCV} 2022 - 17th European Conference, Tel Aviv, Israel, October 23-27, 2022, Proceedings, Part {VIII}}.\hskip 1em plus 0.5em minus 0.4em\relax Springer, 2022, pp. 146--162.

\bibitem{an2025qwen25}
A.~Yang, B.~Yu, C.~Li, D.~Liu, F.~Huang, H.~Huang, J.~Jiang, J.~Tu, J.~Zhang, J.~Zhou, J.~Lin, K.~Dang, K.~Yang, L.~Yu, M.~Li, M.~Sun, Q.~Zhu, R.~Men, T.~He, W.~Xu, W.~Yin, W.~Yu, X.~Qiu, X.~Ren, X.~Yang, Y.~Li, Z.~Xu, and Z.~Zhang, ``Qwen2.5-1m technical report,'' \emph{CoRR}, 2025.

\bibitem{peng2025lvlmeh}
P.~Xu, W.~Shao, K.~Zhang, P.~Gao, S.~Liu, M.~Lei, F.~Meng, S.~Huang, Y.~Qiao, and P.~Luo, ``Lvlm-ehub: {A} comprehensive evaluation benchmark for large vision-language models,'' \emph{{IEEE} Trans. Pattern Anal. Mach. Intell.}, pp. 1877--1893, 2025.

\bibitem{beier2025debias}
B.~Zhu and H.~Zhang, ``Debiasing vision-language models for vision tasks: a survey,'' \emph{Frontiers Comput. Sci.}, p. 191321, 2025.

\bibitem{yao2024effect}
Y.~Jiang, X.~Yan, G.~Ji, K.~Fu, M.~Sun, H.~Xiong, D.~Fan, and F.~S. Khan, ``Effectiveness assessment of recent large vision-language models,'' \emph{Vis. Intell.}, p.~17, 2024.

\bibitem{gen2023cheap}
G.~Luo, Y.~Zhou, T.~Ren, S.~Chen, X.~Sun, and R.~Ji, ``Cheap and quick: Efficient vision-language instruction tuning for large language models,'' in \emph{Advances in Neural Information Processing Systems 36: Annual Conference on Neural Information Processing Systems 2023, NeurIPS 2023, New Orleans, LA, USA, December 10 - 16, 2023}, 2023.

\bibitem{shicheng2025crossm}
S.~Xu, L.~Pang, Y.~Zhu, H.~Shen, and X.~Cheng, ``Cross-modal safety mechanism transfer in large vision-language models,'' in \emph{The Thirteenth International Conference on Learning Representations, {ICLR} 2025, Singapore, April 24-28, 2025}.\hskip 1em plus 0.5em minus 0.4em\relax OpenReview.net, 2025.

\bibitem{jiaxian2023from}
J.~Guo, J.~Li, D.~Li, A.~M.~H. Tiong, B.~Li, D.~Tao, and S.~C.~H. Hoi, ``From images to textual prompts: Zero-shot visual question answering with frozen large language models,'' in \emph{{IEEE/CVF} Conference on Computer Vision and Pattern Recognition, {CVPR} 2023, Vancouver, BC, Canada, June 17-24, 2023}.\hskip 1em plus 0.5em minus 0.4em\relax {IEEE}, 2023, pp. 10\,867--10\,877.

\bibitem{maksim2024vlrm}
M.~Dzabraev, A.~Kunitsyn, and A.~Ivaniuta, ``{VLRM:} vision-language models act as reward models for image captioning,'' \emph{CoRR}, 2024.

\bibitem{zhou2025improving}
Y.~Zhou, L.~Song, and J.~Shen, ``Improving medical large vision-language models with abnormal-aware feedback,'' \emph{arXiv preprint arXiv:2501.01377}, 2025.

\bibitem{zhou2024less}
Y.~Zhou, J.~Zhang, G.~Chen, J.~Shen, and Y.~Cheng, ``Less is more: Vision representation compression for efficient video generation with large language models,'' 2024.

\bibitem{zhou2025draw}
Y.~Zhou, J.~Yuan, and Q.~Wang, ``Draw all your imagine: A holistic benchmark and agent framework for complex instruction-based image generation,'' \emph{arXiv preprint arXiv:2505.24787}, 2025.

\bibitem{wang2025complexbench}
C.~Wang, Y.~Zhou, Q.~Wang, Z.~Wang, and K.~Zhang, ``Complexbench-edit: Benchmarking complex instruction-driven image editing via compositional dependencies,'' \emph{arXiv preprint arXiv:2506.12830}, 2025.

\bibitem{chia2024a}
C.~X. Liang, P.~Tian, C.~H. Yin, Y.~Yua, W.~An{-}Hou, L.~Ming, T.~Wang, Z.~Bi, and M.~Liu, ``A comprehensive survey and guide to multimodal large language models in vision-language tasks,'' \emph{CoRR}, 2024.

\bibitem{jianxing2025genera}
J.~Yu, S.~Wang, H.~Lai, W.~Chen, Y.~Rao, Q.~Su, and J.~Yin, ``Generating commonsense reasoning questions with controllable complexity through multi-step structural composition,'' in \emph{Proceedings of the 31st International Conference on Computational Linguistics, {COLING} 2025, Abu Dhabi, UAE, January 19-24, 2025}.\hskip 1em plus 0.5em minus 0.4em\relax Association for Computational Linguistics, 2025, pp. 2261--2276.

\bibitem{yaoting2025multim}
Y.~Wang, S.~Wu, Y.~Zhang, S.~Yan, Z.~Liu, J.~Luo, and H.~Fei, ``Multimodal chain-of-thought reasoning: {A} comprehensive survey,'' \emph{CoRR}, 2025.

\bibitem{che2024learni}
C.~Zhang, Z.~Xiao, C.~Han, Y.~Lian, and Y.~Fang, ``Learning to check: Unleashing potentials for self-correction in large language models,'' \emph{CoRR}, 2024.

\bibitem{chaojie2024q}
C.~Wang, Y.~Deng, Z.~Lv, Z.~Liang, J.~He, S.~Yan, and B.~An, ``Q*: Improving multi-step reasoning for llms with deliberative planning,'' \emph{CoRR}, 2024.

\bibitem{jiangtong2022from}
J.~Li, L.~Niu, and L.~Zhang, ``From representation to reasoning: Towards both evidence and commonsense reasoning for video question-answering,'' in \emph{{IEEE/CVF} Conference on Computer Vision and Pattern Recognition, {CVPR} 2022, New Orleans, LA, USA, June 18-24, 2022}.\hskip 1em plus 0.5em minus 0.4em\relax {IEEE}, 2022, pp. 21\,241--21\,250.

\bibitem{yi2025score}
Q.~Yi, Y.~He, J.~Wang, X.~Song, S.~Qian, X.~Yuan, M.~Zhang, L.~Sun, K.~Li, K.~Lu \emph{et~al.}, ``Score: Story coherence and retrieval enhancement for ai narratives,'' \emph{arXiv preprint arXiv:2503.23512}, 2025.

\bibitem{zhou2025mam}
Y.~Zhou, L.~Song, and J.~Shen, ``Mam: Modular multi-agent framework for multi-modal medical diagnosis via role-specialized collaboration,'' \emph{arXiv preprint arXiv:2506.19835}, 2025.

\bibitem{unknown1998agents}
\emph{Agents and Multi-Agent Systems Formalisms, Methodologies, and Applications, Based on the AI'97 Workshops on Commonsense Reasoning, Intelligent Agents, and Distributed Artificial Intelligence, Perth, Australia, December 1, 1997}.\hskip 1em plus 0.5em minus 0.4em\relax Springer, 1998.

\bibitem{wang2025insectmamba}
Q.~Wang, C.~Wang, Z.~Lai, and Y.~Zhou, ``Insectmamba: State space model with adaptive composite features for insect recognition,'' in \emph{ICASSP 2025-2025 IEEE International Conference on Acoustics, Speech and Signal Processing (ICASSP)}.\hskip 1em plus 0.5em minus 0.4em\relax IEEE, 2025, pp. 1--5.

\end{thebibliography}
\end{document}